\documentclass[journal]{IEEEtran}
\usepackage{amsmath}
\usepackage{cite}
\usepackage{booktabs}
\usepackage{multirow}

\usepackage{color}
\usepackage[table]{xcolor}

\usepackage[colorlinks,
linkcolor=black,
anchorcolor=black,
citecolor=black]{hyperref}

\ifCLASSINFOpdf
   \usepackage[pdftex]{graphicx}
   \graphicspath{{../pdf/}{../jpeg/}}
   \DeclareGraphicsExtensions{.pdf,.jpeg,.png}
\else
   \usepackage[dvips]{graphicx}
   \graphicspath{{../eps/}}
   \DeclareGraphicsExtensions{.eps}
\fi
\begin{document}
%
\title{Middle-level Fusion for Lightweight RGB-D Salient Object Detection}
%
%
%

\author{Nianchang Huang, ~Qiang Jiao, ~Qiang Zhang*,~Jungong Han*
	\thanks{Nianchang Huang, Qiang Jiao and Qiang Zhang are with Key Laboratory of Electronic Equipment Structure Design, Ministry of Education, Xidian University, Xi’an, Shaanxi 710071, China, and also with Center for Complex Systems, School of Mechano-Electronic Engineering, Xidian University, Xi’an, Shaanxi 710071, China. Email: nchuang@stu.xidian.edu.cn and qzhang@xidian.edu.cn.}
	\thanks{Jungong Han is with Computer Science Department, Aberystwyth University, SY23 3FL, UK. Email: jungonghan77@gmail.com }
	\thanks{*Corresponding authors: Qiang Zhang and Jungong Han.} }	

%
%

\markboth{Journal of \LaTeX\ Class Files,~Vol.~14, No.~8, August~2015}%
{Shell \MakeLowercase{\textit{et al.}}: Bare Demo of IEEEtran.cls for IEEE Journals}
%



\maketitle

\begin{abstract}
Most existing lightweight RGB-D salient object detection (SOD) models are based on two-stream structure or single-stream structure. The former one first uses two sub-networks to extract unimodal features from RGB and depth images, respectively, and then fuses them for SOD. While, the latter one directly extracts multi-modal features from the input RGB-D images and then focuses on exploiting cross-level complementary information. However, two-stream structure based models inevitably require more parameters and single-stream structure based ones cannot well exploit the cross-modal complementary information since they ignore the modality difference.
To address these issues, we propose to employ the middle-level fusion structure for designing lightweight RGB-D SOD model in this paper, which first employs two sub-networks to extract low- and middle-level unimodal features, respectively, and then fuses those extracted middle-level unimodal features for extracting corresponding high-level multi-modal features in the subsequent sub-network. 
Different from existing models, this structure can effectively exploit the cross-modal complementary information and significantly reduce the network's parameters, simultaneously. Therefore, a novel lightweight SOD model is designed, which contains a information-aware multi-modal feature fusion (IMFF) module for effectively capturing the cross-modal complementary information and a lightweight feature-level and decision-level feature fusion (LFDF) module for aggregating the feature-level and the decision-level saliency information in different stages with less parameters. Our proposed model has only 3.9M parameters and runs at 33 FPS. The experimental results on several benchmark datasets verify the effectiveness and superiority of the proposed method over some state-of-the-art methods.

\end{abstract}


%
\IEEEpeerreviewmaketitle

\section{Introduction} \label{sec::I}

\IEEEPARstart{S}{alient} object detection (SOD) aims to detect the most human attractive region in the given images \cite{r1}.  It is an important image pre-processing step for many computer vision tasks, such as image classification \cite{r2}, tracking \cite{r4} and segmentation \cite{r6}.  

So far, many existing SOD methods \cite{r7,r8, r9, r12, r14, r13,r82,r83,r84,r85,r86,r87,r88}  have been presented to detect salient objects from RGB images (\emph{i.e., } RGB SOD) and have achieved significantly progress. However, those models may fail in some challenging scenarios, such as the salient objects sharing similar appearances with the backgrounds and the images with complex backgrounds. Recently, researchers try to introduce the depth images to address those issues, due to the fact that depth images can provide some geometrical information about the scene for complementing the RGB images \cite{r56}. By virtue of the complementary information within RGB and depth (RGB-D) images, RGB-D SOD methods have made significantly progress \cite{r16, r17, r18,    r22,  r24, r25, r26, r27, r28, r53, r54, r55, r70,r73,r75,r76,r77,r78,r79}. 

However, compared with unimodal RGB SOD models, most existing state-of-the-art (SOTA) RGB-D SOD models \cite{r17,r19,r22,r24,r53,r60,r70,r73,r75,r76} require more computational costs and memory consumption to accurately detect the salient objects, since RGB-D SOD models need to process more information from the images of two modalities. This restricts their real-life applications since most of them are running on resource-constrained devices, such as mobiles and on-board computer in car. To address this issue, some lightweight RGB-D SOD models have been presented, which can be roughly divided into two-stream structure based models \cite{r90} and single-stream structure based models  \cite{r69, r67, r77}.
\begin{figure*}[!t]
	\centering
	\includegraphics[width=0.98\textwidth]{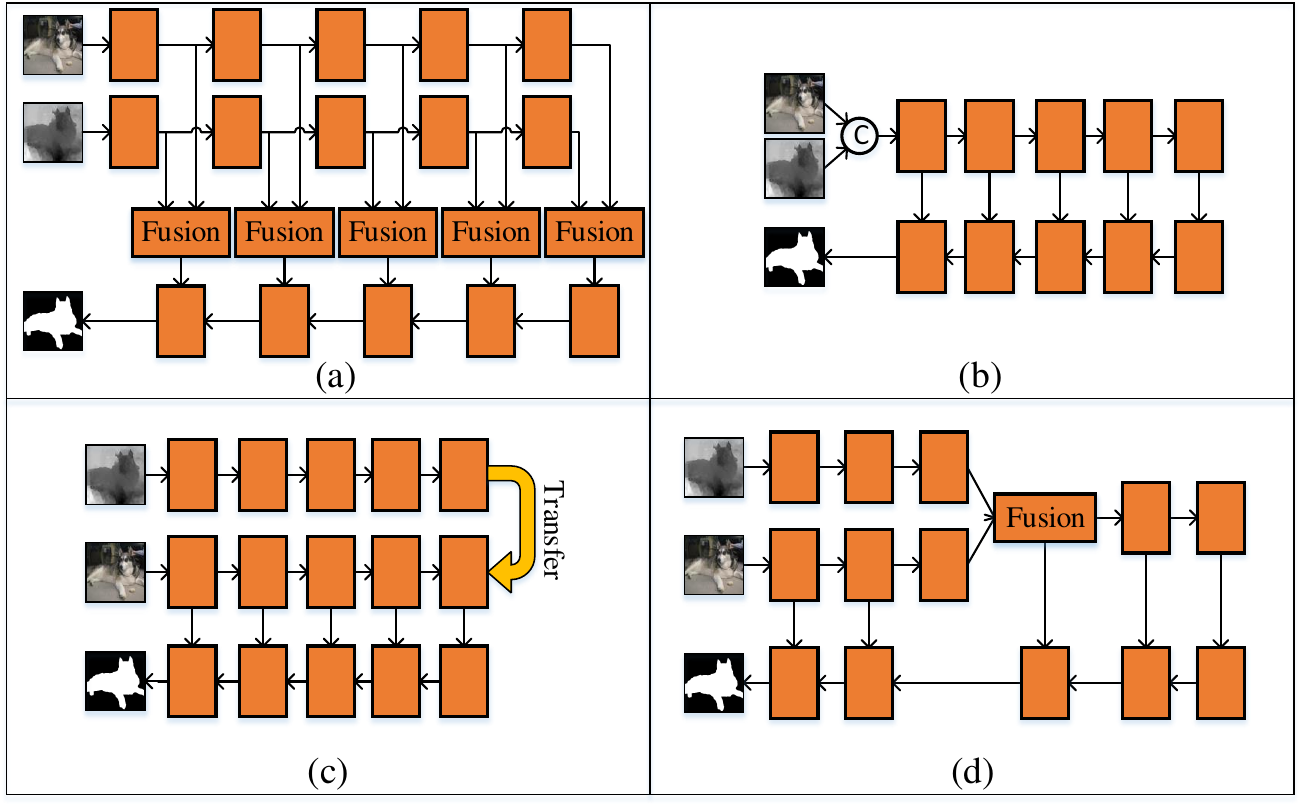}
	\caption{Architectures of most existing RGB-D SOD methods. (a) Two-stream structure. (b) and (c) Single-stream structure. (d) The middle-level fusion structure in our proposed model.}
	\label{fig3}
\end{figure*}

As shown in Fig. \ref{fig3}(a), two-stream structure based models \cite{r90} usually reduce their parameters by designing lightweight unimodal feature extraction, multi-modal feature fusion and saliency prediction modules. For example, ATSA \cite{r90} designed a lightweight asymmetric two-stream architecture, which uses a standard network for RGB images and designs a lightweight depth network (DepthNet) for depth images, to reduce the network's parameters increased by introducing depth images. Then, it proposes a novel depth attention module (DAM) to ensure that the depth features can effectively guide the RGB features by using the discriminative power of depth cues. Considering the modality discrepancy between the RGB and depth images, two-stream structure enables RGB-D SOD models to better exploit the complementary information within the RGB-D images for SOD. However, these models inevitably require more computational costs and memory consumption.

As shown in Fig. \ref{fig3}(b) and (c), compared with two-stream structure based models, single-stream structure based models first reduce their network's parameters by only using one subnetwork for feature extraction. Then, they design different lightweight multi-level feature fusion strategies to effectively exploit the extracted features across different levels for SOD. For example, as in Fig.~\ref{fig3}(b),  DANet \cite{r69} concatenates the RGB and depth images as the four-channel inputs of a SOD model.  As in Fig.~\ref{fig3}(c), inspired by knowledge distillation, A2dele \cite{r67} and CoNet \cite{r77} first transfer the depth knowledge from the depth stream to the RGB stream in the training phrase and then employ the RGB stream only for SOD in the testing phrase. Although, single-stream structure based models can effectively reduce network's parameters, they cannot well exploit the cross-modal complementary information, since they use the shared subnetwork for feature extraction, while ignoring the modality difference between the RGB and depth images. Meanwhile, those depth information cannot be fully transferred into RGB stream and those transfered information cannot fully represent the depth information extracted from depth images, due to their modality discrepancy. As a result, these models still have a large performance gap with other RGB-D SOD models.

As shown in Fig.~\ref{fig3}(d), to solve the limitations of existing lightweight models, we revisit the middle-level feature fusion structure. As shown in Fig.~\ref{fig3}(d), it first employs two subnetworks for extracting low-level unimodal features from the RGB and depth images, respectively, and then designs a multi-modal feature fusion module to fuse the extracted unimodal RGB and depth features. After that, it employs another subnetwork to extract high-level multi-modal features. Finally, it designs a saliency prediction module for deducing the saliency maps. Compared with existing models, the middle-level feature fusion structure has many advantages. First, compared with two-streamed structure, the middle-level feature fusion structure can significantly reduce the network's parameters by (1) using one subnetwork for extracting high-level features since the high-level features contain far more feature channels than those of low-level features; (2) using the multi-modal feature fusion module once. Secondly, compared with single-stream structure, the middle-level feature fusion structure can well extract the cross-modal complementary information from the input RGB-D images, due to the fact that it employs two independent subnetworks for low-level feature extraction and a shared sub-network for multi-modal feature fusion as well as high-level feature extraction. Therefore, in this paper, we will propose a novel middle-level feature fusion structure based lightweight RGB-D SOD model. To the best of our knowledge, this is also the first work which employs middle-level feature fusion structure for lightweight SOD. 


In this model, a novel information-aware multi-modal feature fusion (IMFF) module is first designed to effectively capture the cross-modal complementary information within RGB-D images. The idea behind our proposed IMFF module is that the multi-modal feature fusion aims to exploit all the useful information in the RGB and depth images, including their complementary and redundant ones. The question is that we do not know which local region contains useful information and which one does not in the training and testing phrases. However, the unimodal features from an arbitrary local regions of RGB or depth images can indirectly reflect the amount of information in this region. Generally speaking, a high informative region is more likely to contain more useful information, while a less informative region (\emph{e.g.,} low-quality regions and background regions) may contain more ordinary information. Therefore, the local regions with useful information of the input RGB and depth images may be identified by searching their corresponding informative regions. Meanwhile, compared with RGB SOD, the informative local regions of the input RGB and depth images may be more easily distinguished by comparing their unimodal features from the same local regions in an information-aware feature space. To this end, in our proposed IMFF module, the local unimodal features from the RGB and depth images, respectively, are first project into an information-aware feature space and then the differences in the amount of their contained information are compared to distinguish whether this local region in RGB (depth) image is informative or not.  
 
Then, a novel lightweight feature-level and decision-level feature fusion (LFDF) module is designed to aggregate the feature-level and the decision-level saliency information in different stages with less parameters. In our proposed LFDF module, all the input features are first reduced into 64 channels to reduce our model's parameters. Then, our proposed LFDF module employs a novel feature-level and the decision-level saliency information aggregation structure to significantly aggregate all the information in the features and the saliency maps across different stages. This can also make up the performance degradation caused by parameter reduction. 

In summary, the main contributions of this work are as follows:

(1) By revisiting the middle-level feature fusion, a novel lightweight RGB-D SOD model is presented in this paper, which achieves high efficiency, good accuracy and small model size, thus contributing to SOD's real-life applications. 

(2) A novel information-aware multi-modal feature fusion (IMFF) module is designed to exploit all the discriminative saliency information in the RGB and depth images. Different from most existing models which employ simple fusion strategies (\emph{e.g.,} concatenation and element-wise addition), our proposed IMFF module fuses multi-modal features according to the amount of their contained information.

(3) A novel lightweight feature-level and decision-level feature fusion (LFDF) module is presented to effectively aggregate the feature-level and decision-level saliency information of different stages with less parameters for better saliency prediction.

The rest of this paper are organized as follows: In Section 2, we briefly introduce some previous works related to RGB and RGB-D salient object detection. In Section 3, the details of proposed method are presented, including the architecture and loss function. In Section 4, we conduct a series of experiments to validate proposed model. Finally, in Section 5, a brief conclusion is made for this paper.

\section{Related work} \label{sec::II}

\subsection{RGB SOD}
 
Conventional RGB SOD models mainly integrate different kinds of hand-designed features and prior knowledge together to model the focused attention of human beings \cite{r14, r15, r43}. Recently, convolutional neural networks (CNNs)-based  RGB SOD models \cite{r7,r8, r9, r12, r14, r13,r82,r83,r84,r85,r86,r87,r88}  has dominated this field due to its capability of extracting high-level global information and low-level local details, simultaneously.  Lots of them try to employ the cross-level contextual information for saliency detection. For example, DSSNet \cite{r9} designed an enhanced HED architecture to aggregate the multi-level context information from the deeper layers to the shallower ones with the aid of multiple short connections. Afterwards, some RGB SOD models try to introduce the multi-scale context information to handle the large variants in the shapes and sizes of salient objects. For example, MINet \cite{r84} proposed a self-interaction module to enable the network adaptively extract multi-scale information from the input images. By integrating the self-interaction modules into their saliency prediction module, their network may adaptively deal with scale variation of different samples during the training and testing stages. 

More recently, the edge information from the salient objects has been revisited to address the blurred boundary problem. For example, ENFNet \cite{r85}  proposed an edge guidance block to embed the edge prior knowledge into hierarchical feature maps. The edge guidance block simultaneously performs the feature-wise manipulation and spatial-wise transformation for effective edge embeddings. Besides, the part-object relationships are also exploited for solving the problems that fundamentally hinge on relational inference for visual saliency detection \cite{r87,r88}. For example, both TSPOANet \cite{r88} and TSPORTNet \cite{r87} employed the Capsule Network (CapsNet) to  dig into part-object relationships for SOD. 

\begin{figure*}[!t]
	\centering
	\includegraphics[width=0.95\textwidth]{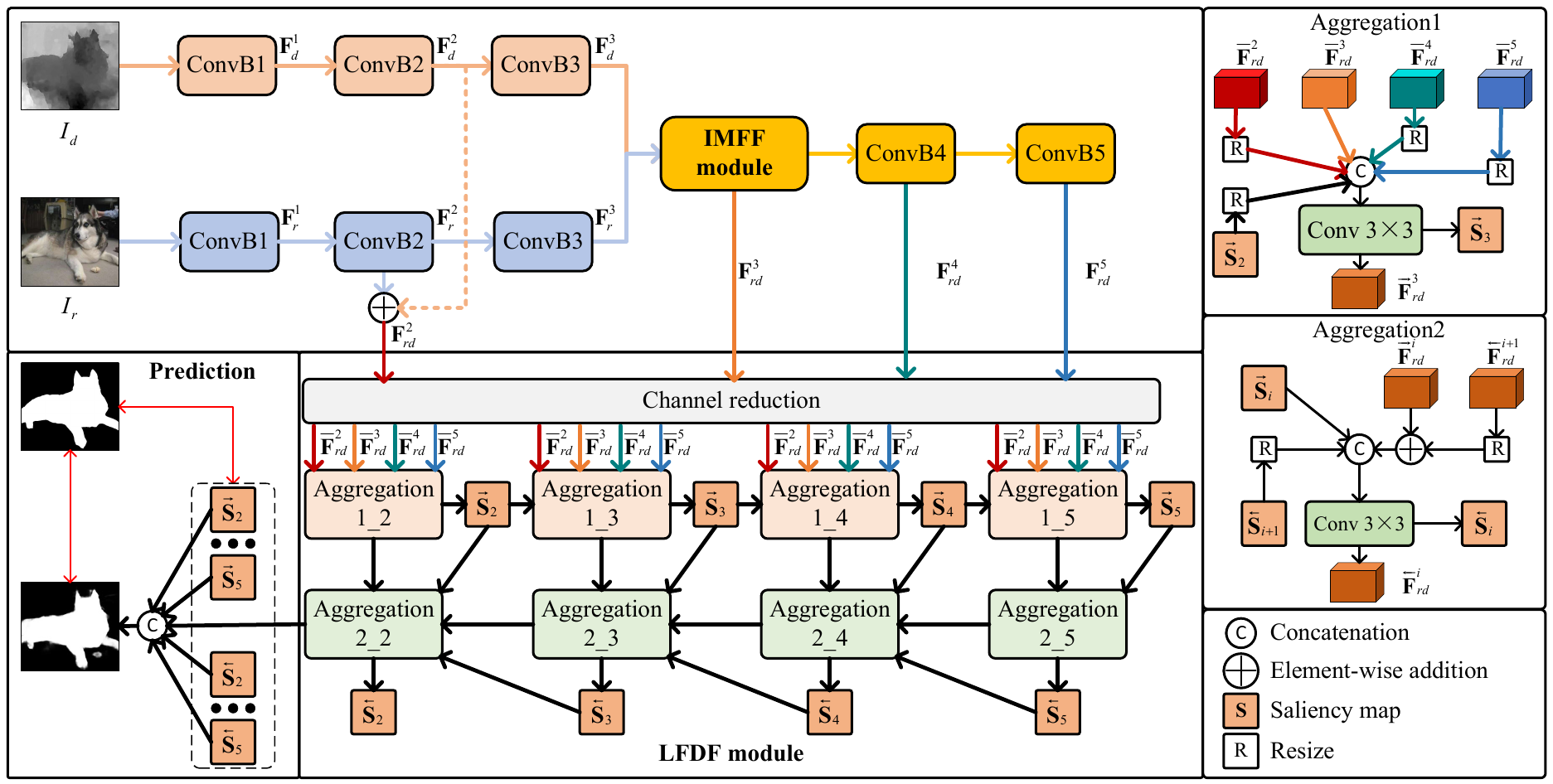}
	\caption{Framework of the proposed lightweight RGB-D SOD model.}
	\label{fig1}
\end{figure*}

\begin{figure*}[!t]
	\centering
	\includegraphics[width=0.95\textwidth]{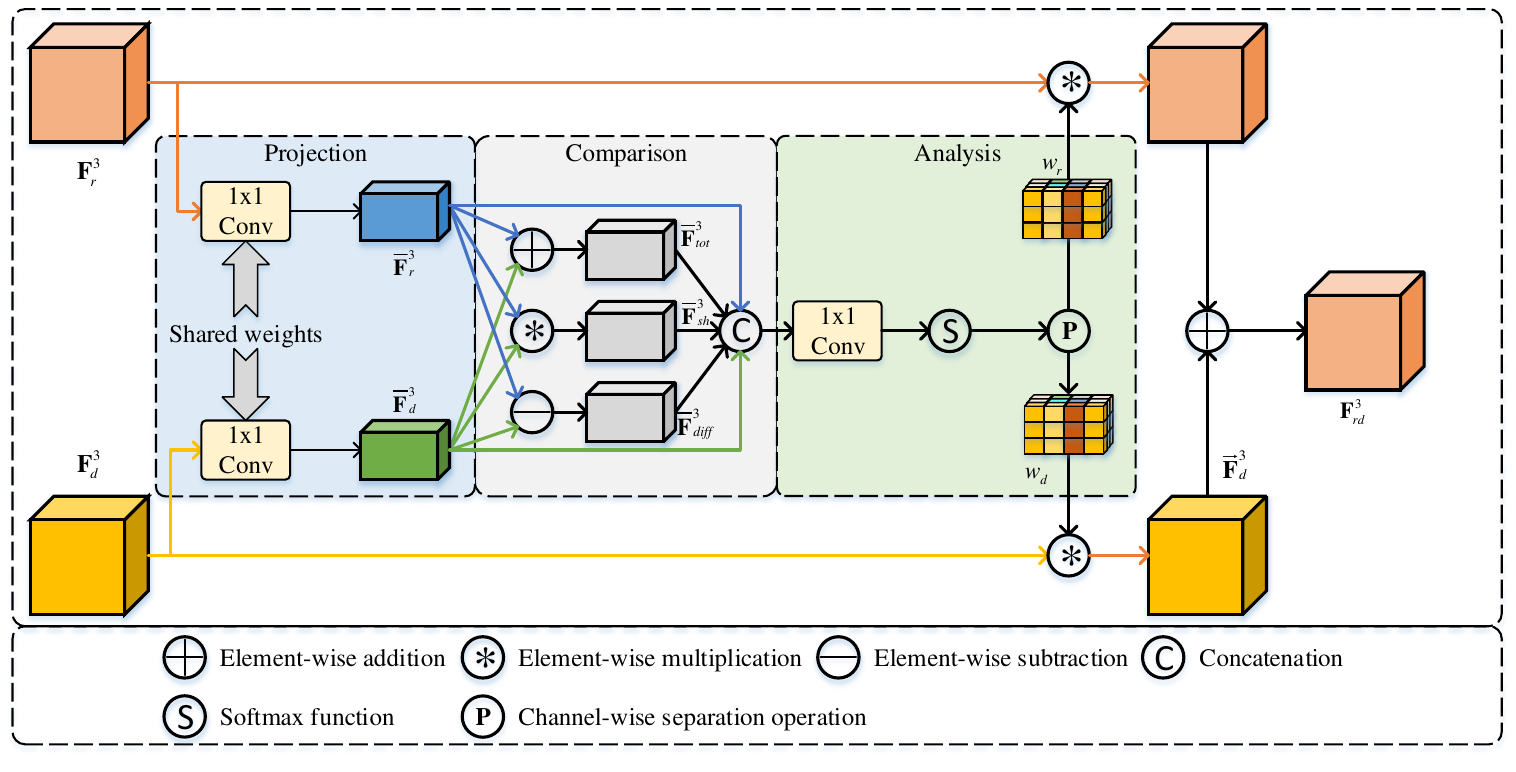}
	\caption{Architectures of our proposed IMFF module.}
	\label{fig2}
\end{figure*}

\subsection{RGB-D SOD}

Recently, RGB-D SOD has received great research interests to exploit the complementary information in RGB-D image for boosting SOD. A complete survey on RGB-D SOD methods is beyond the scope of this paper and we refer the readers to a recent survey paper \cite{r89} for more details. In general, most existing RGB-D SOD models can be summarized into three categories, \emph{i.e.,} pixel-level fusion, feature-level fusion and decision-level fusion. 

Pixel-level fusion based models \cite{r16, r23, r25, r27, r69, r78} directly take the RGB and depth images as the input of four channels for SOD models.  For example, DANet \cite{r69} employed a single stream encoder which concatenates the input RGB and depth images as the four channels of inputs and then designs a depth-enhanced dual attention module to filter the mutual interference between depth prior and appearance prior, thereby enhancing the overall contrast between foreground and background.  

Feature-level fusion based models \cite{r17, r19, r22, r73,r75, r60, r76 } first extract the unimodal RGB and depth features from the input RGB and depth images, respectively, and then fuse them to capture their complementary information for SOD. For example, JCUF \cite{r60} first employed two sub-network to extract unimodal RGB and depth features from the input RGB and depth images, respectively, and then designed a multi-branch feature fusion module to jointly use the fused cross-modal features and the unimodal RGB and depth features for SOD. To effectively capture the cross-modal complementary information within RGB-D images, EBFS \cite{r76} designed a novel multi-modal feature interaction module to simultaneously capture the first-order and the second-order statistical characteristics between the unimodal RGB and depth features. 
 
Decision-level fusion based models \cite{r18, r21, r47} first deduce two saliency maps from the input RGB  and depth images, respectively, and then fuse the two saliency maps by using some well designed weight maps. For example, QAMSOD \cite{r47} first deduced two saliency maps from the input RGB and depth images, respectively, and then fused the two saliency maps by using two weight maps. It should be noted that they generate the two weight maps by using a deep  reinforcement learning algorithm.

Although RGB-D SOD has achieved great progress recently, most existing RGB-D SOD models require high computational costs and memory consumption to obtain high accuracy. Considering that, we propose a novel lightweight RGB-D SOD model in this paper.

\section{Proposed model} \label{sec::III}

As shown in Fig.~\ref{fig1}, the proposed lightweight RGB-D SOD model employs a middle-level feature fusion structure. Given the input RGB images (denoted by $I_r$) and depth images (denoted by $I_d$), two sub-networks are first employed to extract their unimodal features, respectively. As a result, three levels of the unimodal RGB and depth features (denoted by $\mathbf{F}_r^i$ and $\mathbf{F}_d^i$, $i$=1,2,3, respectively) are obtained from the input RGB and depth images, respectively. Then, $\mathbf{F}_r^3$ and $\mathbf{F}_d^3$ are fed into our proposed information-aware multi-modal feature fusion (IMFF) module to exploit their cross-modal complementary information. As a result, the third level of the fused features $\mathbf{F}_{rd}^3$ are obtained. After that, the third level of the fused features are fed into other sub-networks to extract corresponding high-level cross-modal features. Here, two levels of the multi-modal features $\mathbf{F}_{rd}^i, i$=4, 5, are further obtained. Finally, the extracted multi-modal features at different levels are fed into our proposed lightweight feature-level and decision-level feature fusion (LFDF) module for detecting salient objects. It should be noted that, as shown in Fig. \ref{fig1}, the second level of fused features $\mathbf{F}_{rd}^2$ are obtained by  element-wise adding the extracted unimodal features ($\mathbf{F}_r^2$ and $\mathbf{F}_d^2$). Details about these modules will be discussed in the following content.

\subsection{Feature extractors}

As shown in Fig.~\ref{fig1}, there are three feature extractors in our proposed lightweight RGB-D SOD model. Two of them are employed to extract low-level unimodal features from the input RGB and depth images, respectively. While, the rest one is to extract those high-level cross-modal features. Taking advantages from the existing lightweight technologies, we choose the shufflenet \cite{r80} as our feature extractors, which is a classic lightweight classification network. Specifically, the two subnetworks for unimodal feature extraction use  the same structure with the first three convolutional blocks of the shufflenet. Furthermore, their parameters are first pre-trained on ImageNet \cite{r91} and then independently re-trained in our proposed network.  The sub-network for extracting high-level multi-modal features uses the same structure with the last two convolutional blocks of the shufflenet. Its parameters are also pre-trained on Imagenet and re-trained in our model.

\subsection{IMFF module}

As discussed in Section \ref{sec::I}, the proposed IMFF module aims to exploit all the discriminative saliency information in the RGB and depth images for SOD. Based on the assumption that the unimodal features can indirectly reflect those informative and non-informative regions in the input RGB and depth images, the proposed IMFF module tries to fuse the unimodal RGB and depth features from their local regions with abundant amount of information, since those informative regions are more likely to contain discriminative saliency information and those non-informative regions may be low-qualities or backgrounds. Meanwhile, as shown in Fig. \ref{fig1}, our proposed IMFF module are employed only once in our model. Compared with employing multi-modal feature fusion module several times in two-stream structure based models, this can significantly reduce our network's parameters. 
 
\begin{figure}[!t]
	\centering
	\includegraphics[width=0.45\textwidth]{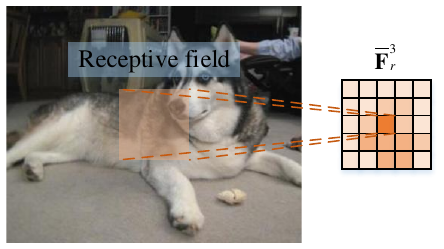}
	\caption{Illustration of the relations between the RGB images and corresponding information-aware features.}
	\label{fig5}
\end{figure}

Specifically, the IMFF module is only employed for the third level of unimodal RGB features $\mathbf{F}_r^3$  and depth features $\mathbf{F}_d^3 \in R^{C_3 \times W_3 \times H_3}$, because we experimentally find that they may have proper receptive field to reflect the informativeness of different local regions in the input images. Here, $C_3$, $W_3$ and $H_3$ are  the feature channels,  width and height of the third level of unimodal RGB or depth features, respectively. As mentioned above, the features in $\mathbf{F}_r^3$ and  $\mathbf{F}_d^3$ can indirectly reflect the the amount of information contained in different local regions in the input RGB and depth images. Considering that, as shown in Fig.~\ref{fig2}, the third level of unimodal RGB and depth features are first projected into an information-aware feature space by using a shared transfer function, \emph{i.e.,} 
\begin{equation}\label{eq1}
\begin{split}
& \overline{\mathbf{F}}_r^3 = \operatorname{Conv}(\mathbf{F}_r^3, \theta_1),  \\& \overline{\mathbf{F}}_d^3 = \operatorname{Conv}(\mathbf{F}_d^3, \theta_1), 
\end{split}
\end{equation} 
where $\operatorname{Conv}(*, \theta_1)$ is a convolutional layer and its parameters $\theta_1$. It serves as the transfer function. $\overline{\mathbf{F}}_r^3$ and $\overline{\mathbf{F}}_d^3 \in R^{C_3 \times W_3 \times H_3}$ denote the projected information-aware features. By doing so, as shown in Fig. \ref{fig5}, each local feature of  $\overline{\mathbf{F}}_r^3$ and $\overline{\mathbf{F}}_d^3$ can reflect the amount of information that are contained in its corresponding local RGB and depth image regions (\emph{i.e.,} its respective field) in this information-aware features space.  

Then, given the information-aware features in the different regions of the RGB and depth images, we can analyse the relative informativeness between the RGB and depth images in different local regions and further determine the informative and non-informative regions of the RGB and depth images. For examples, given an arbitrary local region in the input RGB and depth images, we find that their total amount of information is large, their amount of shared information is little and the amount difference of their information is large by comparing their information-aware features. Accordingly, we may inference that both the local regions of the input RGB and depth images are likely to be informative and there are abundant complementary information between the input RGB and depth images in this local region. 

Therefore, as shown in Fig. \ref{fig2}, we analyse the relations between the information-aware features $\overline{\mathbf{F}}_r^3$ and $\overline{\mathbf{F}}_d^3$ to determine those informative and non-informative regions. Concretely, their total amount of information, their amount of shared information and the amount difference of their information are computed by the following equations, \emph{i.e.,}
\begin{equation}\label{eq2}
\begin{split}
& \overline{\mathbf{F}}_{tot}^3 = \overline{\mathbf{F}}_r^3 + \overline{\mathbf{F}}_d^3, \\ & 
\overline{\mathbf{F}}_{sh}^3 = \overline{\mathbf{F}}_r^3 * \overline{\mathbf{F}}_d^3, \\ & 
\overline{\mathbf{F}}_{diff}^3 = \overline{\mathbf{F}}_r^3 - \overline{\mathbf{F}}_d^3,
\end{split}
\end{equation} 
where the features $\overline{\mathbf{F}}_{tot}^3$ reflect the total amount of information in each local region of the RGB and depth images. The features $\overline{\mathbf{F}}_{sh}^3$ reflect the amount of shared information in each local region of the RGB and depth images. The features $\overline{\mathbf{F}}_{diff}^3$ reflect the amount difference of information in each local region of the RGB and depth images. 

After that, as shown in Fig. \ref{fig2}, the selection weights (\emph{i.e.,} $w_r$ and $w_d \in R^{C_3 \times W_3 \times H_3 }$) for each local regions in the input RGB and depth images are generated by 
\begin{equation}\label{eq3}
\begin{split}
(w_r, w_d) =\operatorname{P}(\operatorname{Conv}(\operatorname{Cat}( \overline{\mathbf{F}}_r^3, \overline{\mathbf{F}}_{tot}^3, \overline{\mathbf{F}}_{sh}^3, \overline{\mathbf{F}}_{diff}^3, \overline{\mathbf{F}}_d^3),  \theta_2)),
\end{split}
\end{equation} 
where $\operatorname{Conv}(*, \theta_2)$ denotes a 1$\times$1 convolutional layer and its parameters $\theta_2$ for fitting the inherent relations between those information-aware features. $\operatorname{Cat}(*)$ denotes the concatenation operation.  $\operatorname{P}(*)$ denotes the channel-wise separation operation. Here, each local weight in $w_r$ and $w_d$ reflect the informativeness of the corresponding local regions in the input RGB and depth images. Meanwhile, for each local region, we use channel-wise selection rather than spatial-wise selection to further select those informative local features and discard those non-informative ones. Finally, given the $w_r$ and $w_d$, the fused features are obtained by 
\begin{equation}\label{eq4}
\begin{split}
\mathbf{F}_{rd}^3 = w_r * \mathbf{F}_r^3 +  w_d * \mathbf{F}_d^3.
\end{split}
\end{equation}

\begin{figure}[!t]
	\centering
	\includegraphics[width=0.5\textwidth]{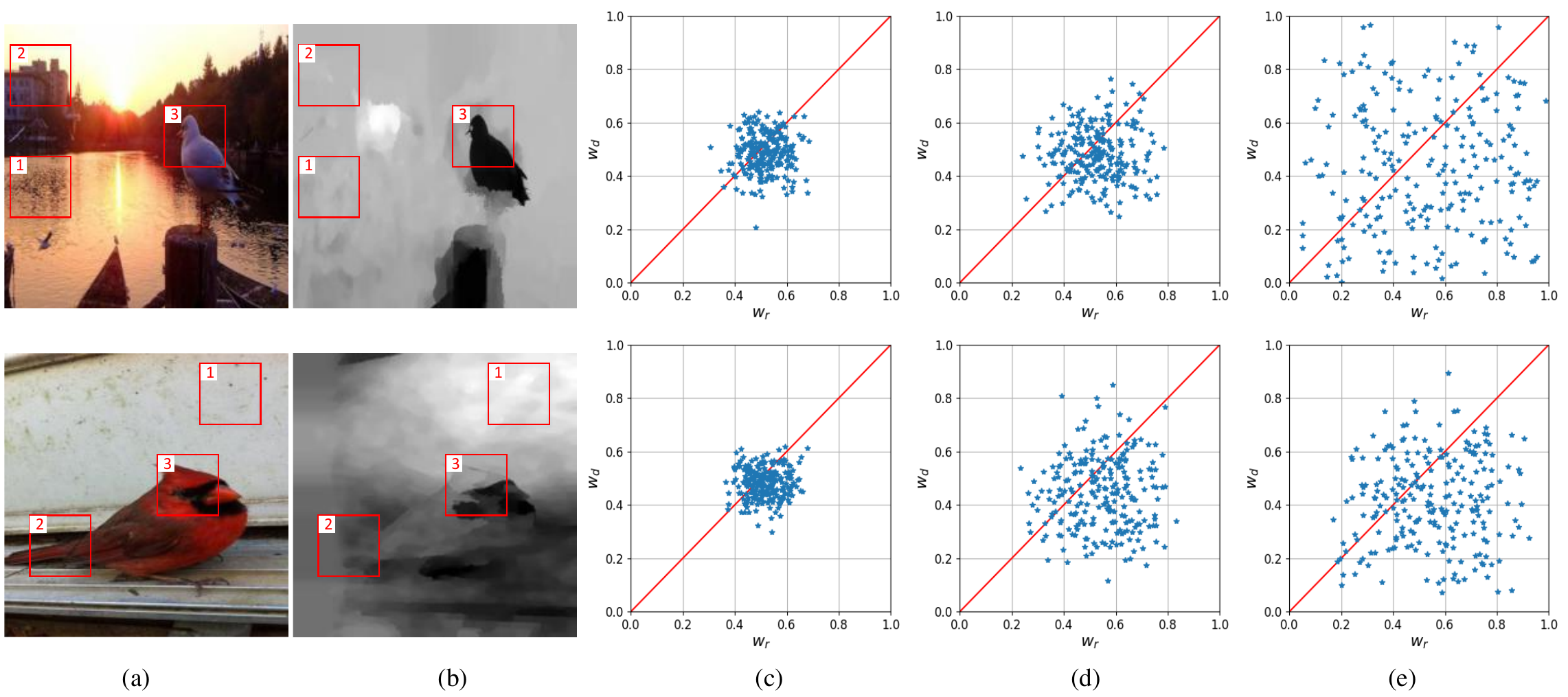}
	\caption{Illustration of some $w_r$ and $w_d$ in different regions of the input RGB and dept images. (a) RGB images. (b) Depth images. (c)-(e) The $w_r$ and $w_d$ in the locations 1, 2 and 3. Here, the horizontal axis is the $w_r$ and the vertical axis is the $w_d$. }
	\label{fig6}
\end{figure}

As shown in Fig. \ref{fig6}, we visualize some $w_r$ and $w_d$ in different regions of the input RGB and depth images, respectively. Generally speaking, the RGB images contain more information than the depth images. Therefore, as shown in Fig. \ref{fig6}(c)-(e), IMFF module aligns more higher weights for unimodal RGB features than that for unimodal depth features, \emph{i.e.,} the points bellow the red line are more than those above the red line, especially for the RGB-D images in the second row since its RGB image's quality is higher than the depth image. Furthermore, the local regions 1 are the non-informative regions for RGB and depth images, the local regions 2 are the informative regions for RGB images and the local regions 3 are the informative regions for RGB and depth images. It can be seen that, for the local region 1, IMFF module may generate equal $w_r$ and $w_d$ with the values of 0.4\textasciitilde0.6. This may result from the fact that the unimodal RGB and depth features in those non-informative local regions have low activation values. For the local regions 2, IMFF module may generate more higher $w_r$ than $w_d$. While, IMFF module has the best feature selection ability for the features in the informative regions. 

Therefore, by virtue of our proposed IMFF module, our model's ability of capturing cross-modal complementary information between the RGB and depth images is significantly improved and, meanwhile, its total parameters are also effectively reduced.

\subsection{LFDF module}

Given the extracted multi-modal features ($\mathbf{F}_{rd}^i, i$=2, 3, 4, 5),  our proposed LFDF module aims to effectively exploit the multi-level complementary information. For that, as shown in Fig. \ref{fig1}, it aggregates not only those saliency information contained in these features of different levels (\emph{i.e.,} feature-level information) but also those saliency information contained in the saliency maps deduced in different levels (\emph{i.e.,} decision-level information) from two directions. This may make up the information losing caused by feature channel reduction.

Specifically, as shown in Fig. \ref{fig1}, the channels of all the multi-modal features $\mathbf{F}_{rd}^i, (i$=2,3,4,5) are first reduced into 64 to decrease the computational complexity and memory in our proposed LFDF module, \emph{i.e.,}
\begin{equation}\label{eq5}
\overline{\mathbf{F}}_{rd}^i = \operatorname{Conv}(\mathbf{F}_{rd}^i, \beta_i),  i=2,3,4,5,
\end{equation} 
where $\operatorname{Conv}(*, \beta_i)$ denotes the convolutional layers with corresponding parameters $\beta_i$. Similarly, all the other features in our proposed LFDF module also set their channel to 64 for reducing the computational complexity and memory. Compared with using their original channels, this can effectively reduce our network's parameters in the subsequent steps.

However, reducing feature's channels will inevitably reduce the amount of salient information contained in the multi-level features, which further drops our model's performance. Our proposed LFDF module tries to address this issue by fully exploiting the feature-level and decision-level information. To this end, as shown in Fig.~\ref{fig1}, it aggregates those complementary information across different levels of features and saliency maps in two directions.

Concretely,  it first aggregates feature-level and decision-level saliency information from the shadow levels to high levels to obtain more discriminative features and more accurate saliency map of current level, \emph{i.e.,}
\begin{equation}\label{eq6}
\overrightarrow{\mathbf{F}}_{rd}^i =\left\{\begin{split}  &
\operatorname{Conv}(\operatorname{Cat}(\overline{\mathbf{F}}_{rd}^2,...,\operatorname{Resize}(\overline{\mathbf{F}}_{rd}^5)), \gamma_i), i=2,
\\ & 
\operatorname{Conv}(\operatorname{Cat}(\operatorname{Resize}(\overline{\mathbf{F}}_{rd}^2),.., \overline{\mathbf{F}}_{rd}^i,..,\operatorname{Resize}(\overline{\mathbf{F}}_{rd}^5), \\ & \ \ \ \ \ \ \ \ \ \ \ \ \ \ \operatorname{Resize}(\overrightarrow{\mathbf{S}}_{i-1}), \gamma_i)),  i=3,4,5.
\end{split}\right.
\end{equation} 
\begin{equation}\label{eq7}
\overrightarrow{\mathbf{S}}_i = \operatorname{Conv}(\overrightarrow{\mathbf{F}}_{rd}^i, \alpha_i),  i= 2,3,4, 5,
\end{equation} 
where  $\mathbf{F}_{rd}^i$ denotes the $i$-th level of the multi-modal features. $\operatorname{Conv}(*,\gamma_i)$ denotes the convolutional layers and its parameters $\gamma_i$. $\operatorname{Cat}(*) $ denotes the concatenation operation. $\operatorname{Resize}(*)$ denotes the bilinear interpolation for resizing the shape of the features from different levels into the shape of the $i$-th level features. $\overrightarrow{\mathbf{S}}_i$ denotes the corresponding saliency maps and $\operatorname{Conv}(*, \alpha_i)$ denotes the 1$\times$1 convolutional layers and its parameters $\alpha_i$ for saliency map prediction.  In this way, the multi-modal features will be re-exploited multi times and the saliency map from previous step will guide our model to better extract high-level semantic information in current stage. 

After that, those feature-level and decision-level saliency information is further aggregated from the high levels to their shadow levels by
\begin{equation}\label{eq8}
\overleftarrow{\mathbf{F}}_{rd}^i =\left\{\begin{split}  &
\operatorname{Conv}(\operatorname{Cat}(\operatorname{Resize}(\overleftarrow{\mathbf{F}}_{rd}^{i+1})+\overrightarrow{\mathbf{F}}_{rd}^{i}, \operatorname{Resize}(\overleftarrow{\mathbf{S}}_{i+1}),  \\ &   \ \ \ \ \ \ \ \ \ \ \ \ \ \  \overrightarrow{\mathbf{S}}_i), \vartheta_i),  i=2,3,4,
\\ & 
\operatorname{Conv}(\operatorname{Cat}(\overrightarrow{\mathbf{F}}_{rd}^5, \overrightarrow{\mathbf{S}}_5), \vartheta_i), i=5.
\end{split}\right.
\end{equation} 
\begin{equation}\label{eq9}
\overleftarrow{\mathbf{S}}_i = \operatorname{Conv}(\overleftarrow{\mathbf{F}}_{rd}^i, \varepsilon_i), i=2,3,4,5,
\end{equation} 
where $\operatorname{Conv}(*,\vartheta_i)$ denotes the convolutional layers and its parameters $\vartheta_i$. $\overleftarrow{\mathbf{S}}_i$ denotes the corresponding saliency maps and $\operatorname{Conv}(*, \varepsilon_i)$ denotes the 1$\times$1 convolutional layers and its parameters $\varepsilon_i$ for saliency map prediction. In this way, the feature-level and decision-level saliency information is effectively aggregated from  high levels to shadow levels. 

Finally, the final saliency map $\mathbf{S_o}$ is obtained by jointly employing the feature-level and decision-level saliency information, \emph{i.e.,}
\begin{equation}\label{eq10}
\begin{split}
\mathbf{S_o} = \operatorname{Conv}(\operatorname{Cat}(& \overleftarrow{\mathbf{F}}_{rd}^2, \overleftarrow{\mathbf{S}}_2,..,\operatorname{Resize}(\overleftarrow{\mathbf{S}}_5),\overrightarrow{\mathbf{S}}_2,.., \\& \operatorname{Resize}(\overrightarrow{\mathbf{S}}_5)),  \iota ), 
\end{split}
\end{equation} 
where $\operatorname{Conv}(*,\iota)$ denotes a 1$\times$1 convolutional layer and its parameters $\iota$.

By virtue of our proposed LFDF module, the computational complexity and memory usage of our proposed model are significantly decreased by reducing the channels of all the features in LFDF module. Meanwhile, its corresponding issue of the amount of information dropping is also well addressed by increasing use efficiency of the feature-level and decision-level saliency information. The experimental results in Section \ref{sec::IV} proves that our proposed LFDF module can make up the performance drop to extents by aggregating the feature-level and decision-level saliency information. 

\subsection{Loss Function}

We employ the cross-Entropy (CE) loss and edge loss to train our proposed network. Among that, CE loss is widely used in the SOD and is expressed by:
\begin{equation}\label{eq_10}
\operatorname{CE}(\mathbf{S}, \mathbf{Y})=\mathbf{Y}\log (\mathbf{S})+(1-\mathbf{Y})\log (1-\mathbf{S}).
\end{equation}
Here, $\mathbf{S}$ denotes corresponding saliency map and $\mathbf{Y}$ denotes the ground truth. And, edge loss is used to refine the boundaries of the generated saliency maps, which is expressed by 
\begin{equation}\label{eq_11}
{{\zeta }_{Ed}}= \operatorname{MSE}(\operatorname{Sobel}(\mathbf{S}), \operatorname{Sobel}(\mathbf{Y})),
\end{equation}
where $\operatorname{MSE}(*)$ denotes the $\operatorname{MSE}$ loss. $\operatorname{Sobel}(*)$ denotes the $\operatorname{Sobel}$ edge detector.

For better training, the CE loss and edge loss are employed for all saliency maps generated in our proposed model, including the final saliency map $\mathbf{S_o}$ and those middle-level saliency maps $\overrightarrow{\mathbf{S}}_i$ and $\overleftarrow{\mathbf{S}}_i, i=2,3,4,5$. Therefore, the overall loss is expressed by
\begin{equation}\label{eq_13}
\begin{split}
{{\zeta }}=& \operatorname{CE} \left( \mathbf{S_o}, \overrightarrow{\mathbf{Y}} \right ) + {{\zeta }_{Ed}} \left( \mathbf{S_o}, \overrightarrow{\mathbf{Y}} \right ) \\& +\sum\limits_{i=2}^{5}{ \left ( \operatorname{CE} \left( \overrightarrow{\mathbf{S}}_i,  \mathbf{Y}  \right )   + {{\zeta }_{Ed}} \left ( \mathbf{S}_{i}, \mathbf{Y}  \right) \right ) } \\ & + \sum\limits_{i=2}^{5}{ \left ( \operatorname{CE} \left( \overleftarrow{\mathbf{S}}_i, \mathbf{Y}_{i} \right ) + {{\zeta }_{Ed}} \left (\mathbf{S}_{i},  \mathbf{Y}  \right) \right ) }.
\end{split}
\end{equation}

\section{Experiments} \label{sec::IV}

\subsection{Datasets}

Our experiments are conducted on four widely used RGB-D SOD datasets: NJU2000 \cite{r26}, NLPR \cite{r49}, STEREO \cite{r50} and SIP \cite{r27}. Among that, NJU2000 dataset \cite{r26} captures and annotates 2003 RGB-D images under diverse objects and complex, challenging scenarios. NLPR dataset \cite{r49} captures and annotates 1000 RGB-D images by using Kinect. It contains a variety of indoor and outdoor scenes under different illumination conditions. STEREO dataset \cite{r50} contains 797 RGB-D images. SIP is a recently proposed dataset, which contains 1000 accurately annotated high-resolution RGB-D images. 

For fair comparisons, we follow the same data split way as  in \cite{r27, r68} and \cite{r75}. Concretely, we randomly sample 1485 RGB-D images from the NJU2K dataset and 700 RGB-D images from the NLPR dataset as our training set. The remaining images in the NJU2K and NLPR datasets and the whole datasets of STEREO and SIP are used for testing.

\subsection{Evaluation Metrics}

We adopt the widely used metrics (\emph{i.e.,} mean F-measure ($F_\beta$) \cite{r26}, mean absolute error (MAE) \cite{r26}, the S-measure ($S_\alpha$) \cite{r51}, the precision-recall (PR) curves \cite{r26} and E-measure $E_\gamma$) \cite{r39}) to verify the proposed model. 

Among that, F-measure is a weighted harmonic mean of $Precision$ and $Recall$, which evaluates the overall performance of a salient object detection model. It is  defined by
\begin{equation}\label{eq14}
F_{\beta} = \frac{(1+\beta^2) \times Precision \times Recall}{\beta^2 \times Precision + Recall}.
\end{equation}
We set $\beta^2=0.3$ as suggested in \cite{r26}. Here, $Precision$ and $Recall$ are computed by comparing the ground truths and the binarized saliency maps under different thresholds.

$MAE$ computes the difference between the saliency map $\mathbf{S}$ and the ground truth $\mathbf{Y}$. Its formulation is expressed by
\begin{equation}\label{eq15}
MAE = \frac{1}{W \times H} \sum_{x=1}^{W} \sum_{y=1}^{H}|\mathbf{S}(x,y)-\mathbf{Y}(x,y)|,
\end{equation}
where $W$ and $H$ are width and height of the saliency map (or ground truth), respectively.

S-measure ($S_{\lambda}$) is recently proposed in \cite{r51} to evaluate the structural similarities between salient map and ground truth. It jointly computes the region-aware (Sr) and object-aware (So) structural similarity as their final structure metric:
\begin{equation}\label{eq16}
S=\alpha*S_o+(1-\alpha)*S_r,
\end{equation}
where $\alpha \in [0,1]$ is the balance parameter and sets to 0.5. More details are seen in \cite{r51}.

E-measure ($E_{\gamma}$) \cite{r39} considers the pixel-level errors and image-level errors by simultaneously capturing global statistics and local pixel matching information, which is formulated by  
	\begin{equation}\label{eq_12}
E_{\gamma}=\frac{1}{W\times H}\sum_{x=1}^{W}\sum_{y=1}^{H}\mathbf{\phi}_{FM}(\mathbf{S}(x,y),\mathbf{Y}(x,y)),
\end{equation}
where $W$ and $H$ are the width and height of saliency maps. $\mathbf{\phi}_{FM}(*)$ is the enhanced alignment matrix whose details are in \cite{r39}.

Besides, we also report the amount of parameters (millions, M) of existing models and their corresponding inference speed (FPS). To obtain inference speed, we first randomly sample 500 RGB-D images from the training set and then resize them into 352$\times$352. After that, we feed them into corresponding models and compute corresponding inference speed.

\subsection{Implementation}
We construct our proposed model by using the Pytorch \cite{r58} toolbox on a NVIDIA 2080Ti GPU. All the parameters, expect those in feature extractors, in our proposed model are initialized by using the Xavier initialization. We use the  SGD with Nesterov momentum to optimize our proposed model. Its learning rate, weight decay and mini-batch size are set as 2e-3, 5e-4 and 4, respectively. Furthermore, its learning rate will decay by a factor of 0.8 in every 20 epochs. All the images are resized into $224 \times 224$ in training phrase

\subsection{Ablation Experiments and Analyses}

In this section, the ablation experiments for each component of our proposed model are performed on NJU2000 to investigate their validities and contributions.

\subsubsection{Multi-modal feature fusion in which level}

To investigate the impact of the multi-modal feature fusion in different levels, several versions of our proposed method (\emph{i.e.,} Input\_F, L1\_F, L2\_F, L3\_F, L4\_F and L5\_F, for short, respectively) are provided for comparison. Specifically, we keep the LFDF module fixed and employ the IMFF module in different levels. Input\_F denotes that the input RGB and depth images are concatenated as four-channel input of our model. L1\_F,.., L5\_F denote that the IMFF module are employed in the features of level1, level2, level3, level4 and level5, respectively.

The quantitative results of these models are shown in Table.~\ref{tab_1}. It can be seen that employing the IMFF module on the features of high levels requires more parameters, due to the fact that there are more feature channels in the features of high levels. Furthermore, the performance of our proposed model first increases by moving the proposed IMFF module from level1 to level3 and then drops by moving the proposed IMFF module from level3 to level5. This may result from the fact that, when applying our proposed IMFF module in one of the first two levels, the corresponding features have relative small receptive field and corresponding extracted information from such small local regions in input RGB and depth images cannot effectively reflect their amount of information .  Furthermore, for those high-level features, their receptive fields are too large to effectively capture the cross-modal complementary information within RGB-D images. Employing the IMFF module on the features of the third level can obtain the balance of our model's size and performance. Therefore, in this paper, we choose the L3\_F as our final model.

\begin{table}[!t]
	\renewcommand{\arraystretch}{1.3}
	\caption{quantitative results of different versions of the proposed Bi-MCFF module.}
	\label{tab_1}
	\centering
	\begin{tabular}{cccccc}
		\hline	
		Methods	   &Params $\downarrow$  &$MAE\downarrow$   &$F_\beta\uparrow$      &$S_{\lambda}\uparrow$     &$E{\gamma}\uparrow$ \\
		\hline
		Input\_F  & 3.443 &0.062  &0.838 &0.878  &0.893    \\
		L1\_F	  & 3.444 &0.050 &0.868 &0.889 &0.913       \\
		L2\_F	  & 3.518 &0.046 &0.878 &0.893 &0.922   \\
		L3\_F     & 3.893 &0.042 &0.885  &0.898  &0.925     \\
		L4\_F     & 4.920 &0.043&0.878 &0.896& 0.927    \\
		L5\_F     & 8.107 &0.045 &0.879 &0.896 &0.924   \\
		\hline
	\end{tabular}
\end{table}

\subsubsection{Ablation experiments for each module}

We then investigate the impact of each component of our proposed model. Specifically, the `Baseline' model denotes the one that has removed the IMFF module and LFDF module from our proposed lightweight RGB-D SOD model. As shown in Table \ref{tab_2}, both the IMFF module and  LFDF module (\emph{i.e.,} `Baseline+IMFF' and  `Baseline+LFDF') can improve the performance of RGB-D SOD. This verifies that the proposed IMFF module can effectively capture the cross-modal complementary information and the  proposed LFDF module can well exploit the cross-level complementary information. Furthermore, with the collaboration of the IMFF module and LFDF module, our proposed model (\emph{i.e.,} `Baseline+IMFF+LFDF') obtains the best performance. 

\begin{table}[!t]
	\renewcommand{\arraystretch}{1.3}
	\caption{quantitative results of different versions of the proposed Bi-MCFF module.}
	\label{tab_2}
	\centering
	\begin{tabular}{ccccc}
		\hline	
		Methods	    &$MAE\downarrow$   &$F_\beta\uparrow$      &$S_{\lambda}\uparrow$     &$E{\gamma}\uparrow$  \\
		\hline
		Baseline   &0.055 &0.856 &0.884 &0.909      \\
		+IMFF	  &0.051 &0.863 &0.887 &0.914     \\
		+LFDF	 &0.047 &0.872 &0.889 &0.917    \\
		+IMFF+LFDF   &0.042 &0.885  &0.898  &0.925   \\
		\hline
	\end{tabular}
\end{table}

\begin{table*}[!t]
	\renewcommand{\arraystretch}{1.3}
	\caption{quantitative results of different models. }
	\label{tab_4}
	\centering
	\resizebox{\textwidth}{!}{
		\begin{tabular}{c|cccc|cccc|cccc|cccc}
	\hline	
	\multirow{2}*{Datasets}  & \multicolumn{4}{c|}{NJU2000} & \multicolumn{4}{c|}{NLPR}& \multicolumn{4}{c|}{STEREO} & \multicolumn{4}{c}{SIP} \\
	\cline{2-17}
	 &$MAE\downarrow$   &$F_\beta\uparrow$      &$S_{\lambda}\uparrow$     &$E{\gamma}\uparrow$  &$MAE\downarrow$   &$F_\beta\uparrow$      &$S_{\lambda}\uparrow$     &$E{\gamma}\uparrow$  &$MAE\downarrow$   &$F_\beta\uparrow$      &$S_{\lambda}\uparrow$     &$E{\gamma}\uparrow$  &$MAE\downarrow$   &$F_\beta\uparrow$      &$S_{\lambda}\uparrow$     &$E{\gamma}\uparrow$  \\
\hline	
PCA \cite{r17}  &0.059&0.839	&0.876&0.895&0.043&0.802&0.873&0.887&0.063&0.818&0.874&0.887&0.070&0.814&0.842&0.878\\
TSAA \cite{r22} &0.060&0.841&0.879&0.895&0.041&0.819&0.886&0.901&0.059&0.827&0.871&0.893&0.075&0.803&0.834&0.870\\
CPFP \cite{r24} &0.053&0.850&0.895&0.910&0.035&0.840&0.888&0.917&0.051&0.841&0.879 &0.912&0.063&0.820&0.850&0.893\\
D3Net\cite{r27} &0.051 &0.860  &0.895  &0.912  &0.033  &0.852  &0.905  &0.923  &0.048  &0.844  &0.890  &0.908  &0.062  &0.832  &0.864  &0.882  \\
JCUF\cite{r60}  &0.041 &0.881  &0.901  &0.926 &0.030 &0.885  &0.900  &0.932 &0.045 &0.870  &0.895  &0.924 &0.056 &0.854  &0.873  &0.904   \\
ICNet\cite{r54} &0.052 &0.869  &0.894  &0.913 &0.029 &0.884  &0.926  &0.939 &0.044 &0.869  &0.902  &0.925&0.068&0.834&0.853&0.890   \\
ASIFN\cite{r53} &0.047 &0.881  &0.901  &0.926 &0.030 &0.885  &0.900  &0.932 &0.045 &0.870  &0.895  &0.924&-&-&-&-   \\
UCNet\cite{r74} &0.044 &0.885  &0.896  &0.930  &0.025  &0.890  &0.919  &0.950 &0.039  &0.884  &0.902  &0.938 &0.052  &0.867  &0.875  &0.914   \\
DMRA\cite{r68}  &0.050 &0.873  &0.885  &0.919 &0.031 &0.864  &0.898  &0.939 &0.048 &0.867  &0.885  &0.930 &0.085 &0.819  &0.805  &0.843   \\
SSF\cite{r75}   &0.043 &0.884  &0.897  &0.927 &0.026 &0.881  &0.913  &0.946 &0.045 &0.877  &0.892  &0.928 &-&-&-&-   \\
JLDCF\cite{r78} &0.041 &0.884  &0.902  &0.934 &0.022 &0.893  &0.925  &0.954 &0.041 &0.873  &0.902  &0.935 &0.050 &0.873  &0.880  &0.918   \\
BBSNet\cite{r76} &0.039 &0.897  &0.915  &0.934 & 0.026& 0.889 & 0.922 &0.946 &0.046 & 0.866 & 0.895 &0.922 &0.563 &0.855  &0.874  &0.908   \\
BIANet\cite{r79} &0.039 &0.903  &0.915  &0.934 &0.025 &0.894  &0.925  &0.948 &0.044 &0.879  &0.903  &0.925 &0.053 &0.873  &0.882  &0.912   \\
EBFS\cite{r73}   &0.038 &0.895  &0.907  &0.936 &0.028 &0.887  &0.909  &0.940 &0.041 &0.873  &0.900  &0.926 &0.052 &0.863  &0.877  &0.911   \\
   	\hline	
A2dele*\cite{r67} &0.050 &0.869  &0.868  &0.912 &0.029 &0.875  &0.895  &0.940 &0.043 &0.879  &0.884  &0.930 &- &-  &-  &-   \\
DANet* \cite{r69}  &0.047 &0.874  &0.897  &0.920 &0.030 &0.871  &0.908  &0.933 &0.047 &0.857  &0.892  &0.914&0.053 &0.864  &0.876  &0.910   \\
CoNet* \cite{r77}  &0.047 &0.874  &0.894  &0.94 &0.030 &0.864  &0.907  &0.933 &0.041 &0.885  &0.907  &0.937&0.063 &0.844  &0.857  &0.901   \\
OUR-VGG16         &0.035 &0.900  &0.913  &0.941  &0.025 &0.898  &0.923  &0.952  &0.039  & 0.881  &0.903  &0.936  &0.047 &0.880  &0.881  &0.922  \\
OUR-ShuffeNet    &0.042 &0.885  &0.898  &0.925  &0.027 &0.887  &0.917  &0.943  &0.046  & 0.861  &0.885  &0.915  &0.048 &0.871  &0.882  &0.919  \\
\hline
	\end{tabular}}
\end{table*}

\begin{table*}[!t]
	\renewcommand{\arraystretch}{1.3}
	\caption{The model sizes and inference speed of different methods.}
	\label{tab_5}
	\centering
	\resizebox{\textwidth}{!}{
	\begin{tabular}{cccccccccccccc}
		\hline	
Methods	    &JCUF\cite{r60}& UCNet\cite{r74}&DMRA\cite{r68}&SSF\cite{r75}&BBSNet\cite{r76}&BIANet\cite{r79}&EBFS\cite{r73}&A2dele\cite{r67} &DANet \cite{r69}&CoNet \cite{r77}  &ATSA \cite{r90} &OUR-VGG16  & OUR-ShuffeNet  \\   
\hline 
Params (M) $\downarrow$  &94.6          &31.2            &59.7          &33           &49.8            &49.7            &122.7         &15.1             &26.7            &43.7       & 32.2      &18.4 &3.9 \\
Speed (FPS)  $\uparrow$  &8             &24              &20            &11           &26              &22              &7             &69               &32              &34          & 41     &38 &33  \\
		\hline
	\end{tabular}}
\end{table*}


\subsection{Compare with the State-of-the-art}

The proposed model and some existing state-of-the-art RGB-D salient object detection models are evaluated on four benchmark datasets: NJU2000 \cite{r26}, NLPR \cite{r49}, STEREO \cite{r50} and SIP \cite{r27}. The following state-of-the-art models include PCA \cite{r17}, TSAA \cite{r22}, CPFP \cite{r24}, D3Net\cite{r27}, JCUF\cite{r60}, ICNet\cite{r54}, ASIFN\cite{r53}, UCNet\cite{r74}, DMRA\cite{r68}, SSF\cite{r75}, JLDCF\cite{r78}, BBSNet\cite{r76}, BIANet\cite{r79}, EBFS\cite{r73}, A2dele\cite{r67}, DANet \cite{r69} and CoNet \cite{r77}. Among that, A2dele\cite{r67}, DANet \cite{r69} and CoNet \cite{r77} are lightweight models. For fair comparisons, the saliency maps deduced by these existing models are provided by their authors and are tested by our evaluating code. Here, `OUR-VGG16' denotes that the feature extractors of our proposed model employ the structure of VGG16 network \cite{r10} rather than ShuffeNet. `OUR-ShuffeNet' denotes the feature extractors of our proposed model employ the structure of ShuffeNet \cite{r80}. 

\subsubsection{Quantitative analysis}

Table \ref{tab_4} shows the quantitative results of state-of-the-art models. For the NJU2000 dataset, the proposed `OUR-VGG16' achieves the best performance in the metrics of $MAE$ and $E_\gamma$ and obtains competitive results with respect to $F_\beta$ and  $S_\lambda$. For the NLPR and STEREO datasets, our proposed `OUR-VGG16' achieves competitive results with respect to other state-of-the-art models. For the SIP dataset, our proposed `OUR-VGG16' obtains the best performance in all of the metrics. Furthermore, compared with existing state-of-the-art models, `OUR-ShuffeNet' achieves competitive results in all of the metrics. And, compared with existing lightweight RGB-D SOD models, `OUR-ShuffeNet' obtains the best performance with least parameters. 

Furthermore, as shown in Table.~\ref{tab_5}, compared with existing models,  `OUR-VGG16' has 18.4 million parameters, which is the most lightweight and the most fast RGB-D SOD models, expect for A2dele\cite{r67} and ATSA \cite{r90}. While, `OUR-ShuffeNet'  has only 3.9 million parameters. However, compared with `OUR-VGG16', it inference speed is degraded, due to its high structural complexity.  `OUR-ShuffeNet' can still run in most real-time applications. 

\subsubsection{Qualitative  analysis}

\begin{figure*} 
	\centering
	\includegraphics[width=\textwidth, height=0.4\linewidth]{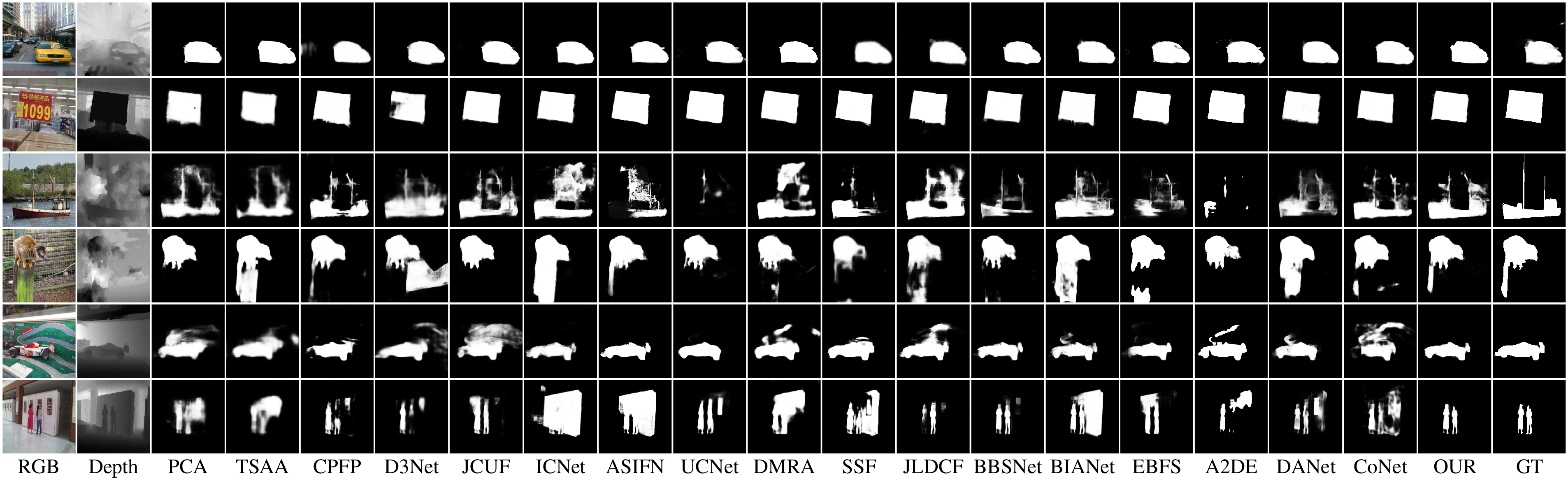}
	\caption{Visualization of saliency maps generated by different models.}
	\label{fig_4}
\end{figure*}

Visualization results under different scenarios are illustrated in Fig. \ref{fig_4}. As shown in the first two rows of Fig. \ref{fig_4}, for the images under some simple scenes, most state-of-the-art methods accurately detect the salient objects. Furthermore, as shown in the last four rows of Fig.~\ref{fig_4}, for those relatively complex scenes, our proposed model can obtain competitive and even better results than those state-of-the-art models. This further verify the effectiveness of our proposed models.

\section{Conclusion}

In this paper, we proposed the first middle-level fusion structure based lightweight RGB-D SOD model. By revisiting the middle-level fusion structure, the proposed model significantly reduces the network's parameters. Furthermore, the proposed IMFF module can effectively capture the cross-modal complementary information with less parameters by exploiting the amount of the information from different local regions in the RGB and depth images. And, the proposed LFDF module can effectively extract the cross-level complementary information by jointly fusing the feature-level and decision-level information cross levels. Based on the middle-level fusion structure, our proposed model has only 3.9M parameters and runs at 33 FPS. Furthermore, experimental results on several benchmarks show that, by virtue of the proposed IMFF and LFDF modules, our proposed model can make up the performance drop caused by reducing parameters to some extents.  

\section*{Acknowledgment}
This work is supported by the National Natural Science Foundation of China under Grant No. 61773301 and 61876140. It is also supported by the Fundamental Research Funds for the Central Universities and the Innovation Fund of Xidian University.

\ifCLASSOPTIONcaptionsoff
  \newpage
\fi




\bibliographystyle{IEEEtran}
\bibliography{work37_ref}

\end{document}